\documentclass[letterpaper, 10 pt, conference]{ieeeconf}  

\IEEEoverridecommandlockouts                              

\usepackage{cite}
\usepackage{graphics} 
\usepackage{epsfig} 
\usepackage{amsmath} 
\usepackage{bm}
\usepackage{amssymb}  
\usepackage{algorithm}
\usepackage{algorithmic}
\usepackage{acronym}
\usepackage{hyperref}
\usepackage{xcolor}

\usepackage{pifont}
\usepackage{multirow}
\usepackage{tikz}
\usepackage{booktabs}
\title{\LARGE \bf
Humanoid Dodgeball: Learning Fast Dynamic Obstacle Avoidance Motions for Humanoids
}
\author{Aoqian~Zhang\textsuperscript{1,*}, Chunzheng~Wang\textsuperscript{1,*}, Haoran~Chen\textsuperscript{1,*}, Annan~Tang\textsuperscript{2},\\
Cheng~Xiang\textsuperscript{3}, Fan~Shi\textsuperscript{1}, and Shuzhi~Sam~Ge\textsuperscript{1}\thanks{\textsuperscript{*}Equal contribution.}\thanks{\textsuperscript{1}Department of Electrical and Computer Engineering, National University of Singapore, Singapore.}\thanks{\textsuperscript{2}Shenzhen EngineAI Robotics Technology Co., Ltd., China.}\thanks{\textsuperscript{3}School of Information and Communication Engineering, Dalian University of Technology, China.}
}
\begin{document}

\maketitle

\suppressfloats[t]

\begin{figure}[t]
\centering
\includegraphics[width=0.8\linewidth]{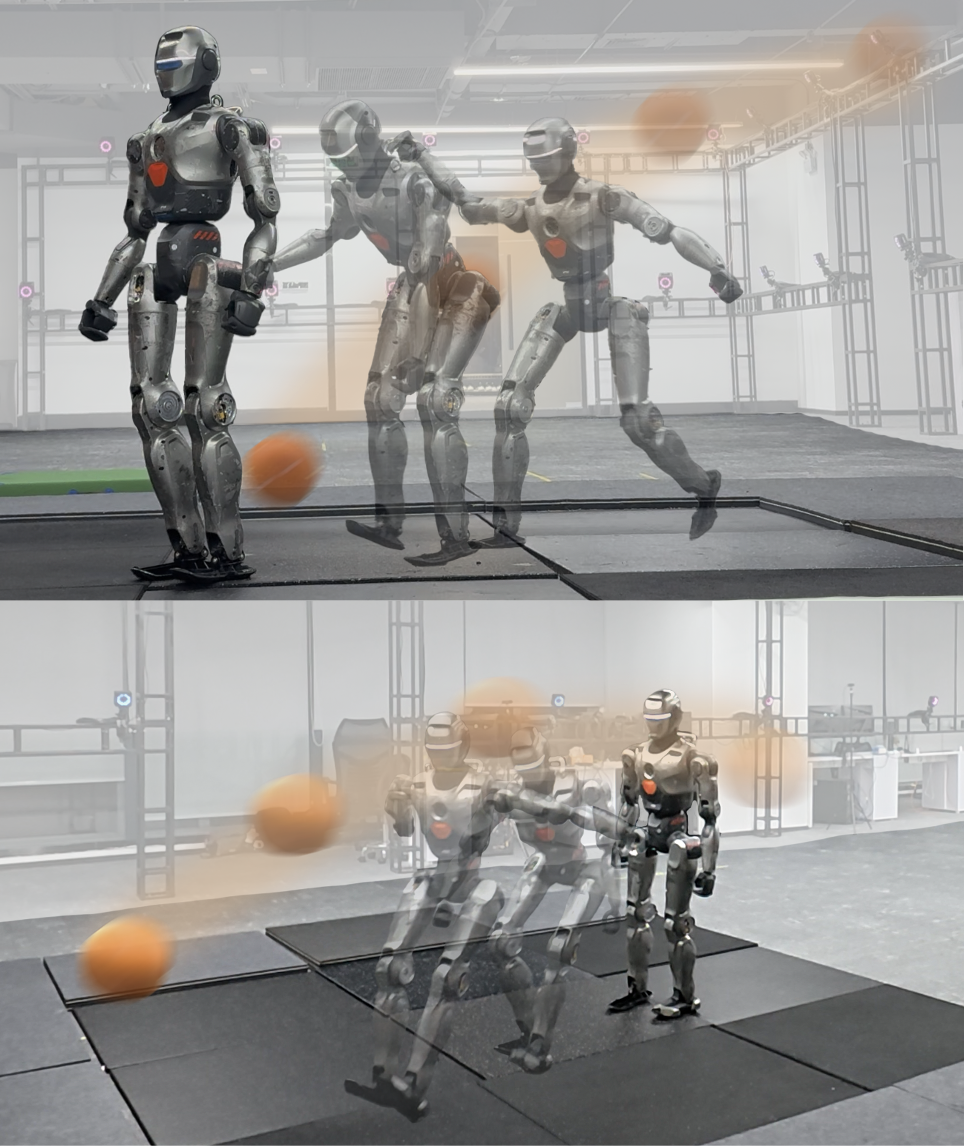}
\caption{Our dynamic obstacle avoidance policy deployed on the EngineAI PM01 humanoid. (Top) The robot dodges by stepping forward. (Bottom) The robot dodges by jumping backward.}
\label{fig:summary}
\vspace{-1.0em}  
\end{figure}

\begin{abstract}
Humanoids have become increasingly agile and capable in real-world environments. To operate safely in dynamic scenarios, they must rapidly evade fast-moving obstacles while maintaining stability and physically feasible whole-body motion. Learning such behaviors remains challenging due to the high-dimensional dynamics of humanoid systems and the need to jointly balance reactive obstacle avoidance, motion naturalness, and control robustness. In this work, we present \textit{Humanoid Dodgeball}, a reinforcement learning (RL) framework for agile dynamic obstacle avoidance in humanoids. Our method enables highly dynamic whole-body evasive maneuvers against incoming projectiles through threat-aware reward shaping, while multiple direction-specific adversarial motion priors regularize the learned behaviors toward stable and diverse human-like motions. Extensive simulation and hardware experiments demonstrate that the proposed framework substantially improves dynamic obstacle avoidance performance across diverse obstacle speeds and launch distances, while preserving robust and physically plausible humanoid motions. Supplementary videos are available at {\hypersetup{hidelinks}\href{https://humanoid-dodge.github.io/}{\textcolor{orange}{\nolinkurl{https://humanoid-dodge.github.io/}}}}.
\end{abstract}

\section{INTRODUCTION}
Recent advances in reinforcement learning for humanoids have significantly improved humanoid whole-body motion tracking and control through training on large human motion datasets \cite{he2025asap,luo2025sonic,ze2026twist2}. These developments have enabled humanoids to acquire a wide range of dynamic and human-like skills, including athletic sports behaviors such as soccer \cite{wang2026vision} and tennis \cite{huang2026tennis}, highly dynamic parkour locomotion \cite{wu2026perceptive}, and complex whole-body tasks such as skateboarding \cite{han2026husky}. To further improve the safety and robustness of humanoid deployment in complex and human-centered environments, recent research has explored protective falling strategies that reduce damage to both the robot and its surroundings during unavoidable failures \cite{meng2025safefall}, as well as collision-free navigation and traversal in cluttered environments with static obstacles \cite{xue2026collisionfree}. These advances represent important steps toward deploying humanoids in real-world environments where safety, robustness, and physical interaction are critical.

Despite these remarkable advances, humanoids still lack the ability to safely and robustly react to fast-moving dynamic objects. The capability to evade incoming obstacles is a fundamental and natural behavior observed in legged animals, including humans, and is essential for agile operation in real-world settings. However, learning such reactive whole-body avoidance behaviors for humanoids remains highly challenging due to the need for rapid reaction, dynamic balance maintenance, and coordinated whole-body motion generation. 

To address these challenges, we present \textit{Humanoid Dodgeball}, a novel humanoid whole-body control framework for dynamic obstacle avoidance. The main contributions of this work are as follows:

\begin{itemize}
\item We design a reinforcement learning framework that enables humanoids to evade fast-moving projectiles through coordinated whole-body evasive maneuvers while maintaining stability.
\item We combine threat-aware rewards, direction-specific motion priors, and capture-point guidance to promote effective avoidance, diverse natural motions, and stable recovery.
\item We validate our proposed method through simulation experiments on two humanoid platforms and hardware experiments on the EngineAI PM01 humanoid.
\end{itemize}

\section{RELATED WORK}
Dynamic obstacle avoidance has been widely studied across different robotic platforms. In aerial robotics, Falanga et al. proposed an event-camera-based framework for quadrotor obstacle avoidance that leverages the microsecond-level response of event cameras to generate low-latency reactive avoidance behaviors using artificial potential fields \cite{falanga2020dynamic}. More recently, Fan et al. introduced an end-to-end reinforcement learning framework that uses multi-frame LiDAR observations to directly generate control commands for avoiding high-speed dynamic obstacles \cite{fan2025flying}. In quadrupedal robots, He et al. proposed Agile But Safe (ABS), a dual-policy reinforcement learning framework that combines agile locomotion and recovery policies through a learned reach-avoid value network for high-speed collision-free navigation \cite{he2024agile}. Wang et al. further proposed Omni-Perception, an end-to-end reinforcement learning framework that directly processes raw LiDAR point clouds using a hierarchical perception network for omnidirectional obstacle avoidance \cite{wang2025omniperception}. To address the slow response of navigation-based avoidance strategies, Xu et al. proposed the Reflexive Evasion Robot (REBot) framework, which introduces a biologically inspired finite-state-machine architecture for reflexive evasion behaviors \cite{xu2025rebot}. REBot enables rapid obstacle avoidance through shorter reaction times and faster whole-body quadruped maneuvers.

Dynamic obstacle avoidance is significantly more challenging for humanoids than for aerial and quadruped robots due to humanoids' high-dimensional whole-body dynamics, ground contact limitations, and strict balance constraints. When faced with fast-moving incoming obstacles, humanoids cannot simply evade threats through low-dimensional velocity modulation or direct body translation. Instead, they must coordinate each body part and foot placement in a fully integrated manner to perform agile evasive maneuvers while maintaining locomotion stability and physically plausible motion. Concurrently with our work, Yang et al. proposed a perception-aware reinforcement learning framework for humanoid dynamic obstacle avoidance that combines control barrier function (CBF)-inspired reward shaping with adversarial motion priors (AMP)-based motion regularization \cite{yang2026pacman}. However, it considers a limited set of obstacle trajectories, which limits the policy's ability to generalize to more diverse projectile conditions. Compared with existing approaches, our method enables more effective dynamic obstacle avoidance for humanoids by achieving higher avoidance success rates and robust evasive behaviors against obstacles approaching from diverse directions and speeds.

\section{METHODS}

\begin{figure*}[h]
    \vspace{8pt}
    \centering
    \includegraphics[width=1.0\linewidth]{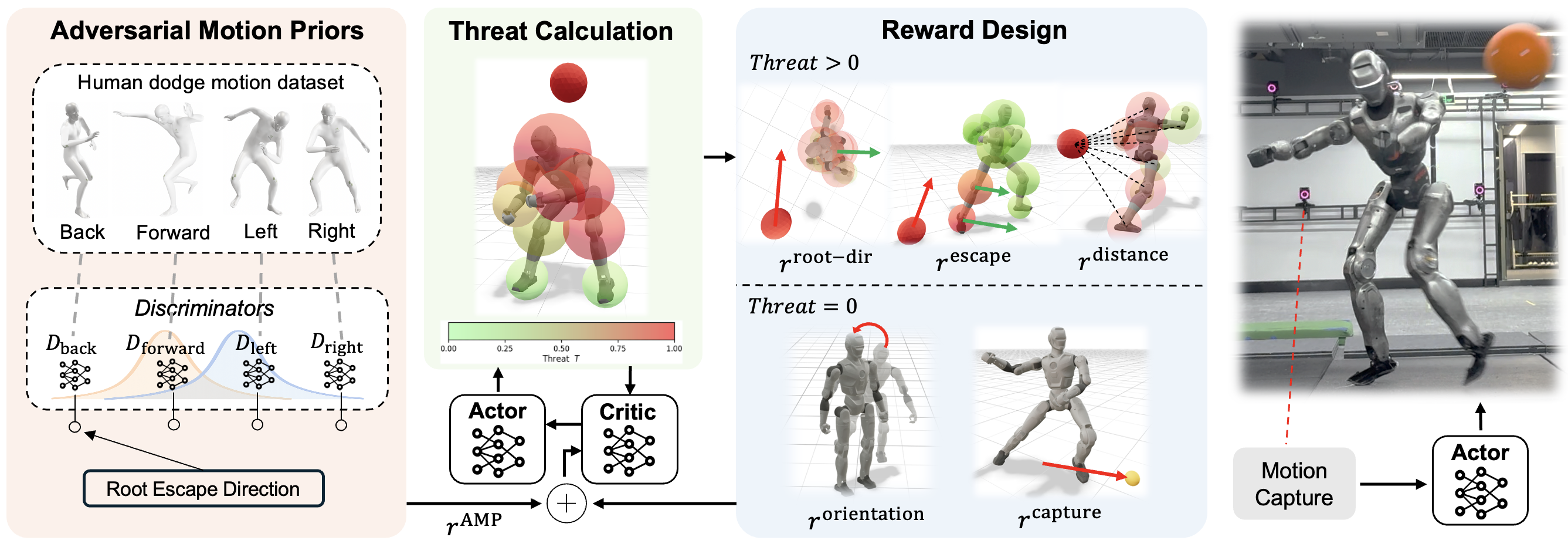}
    \caption{Overview of \textit{Humanoid Dodgeball}. During simulation training, predictive and near-field threat for key bodies are computed to characterize incoming projectile hazards and guide threat-aware reward shaping for whole-body evasion. Direction-specific adversarial motion priors, selected according to the root escape direction, regularize the learned behaviors toward diverse and natural human-like dodge motions. When the threat subsides, recovery rewards, including orientation and capture-point objectives, encourage rapid stabilization. During hardware deployment, the trained actor is executed using obstacle positions estimated from the motion-capture system.}
    \label{fig:pipeline}
\vspace{-1.0em}   
\end{figure*}

\subsection{Problem formulation}
We formulate dynamic obstacle avoidance for humanoid robots as a
discrete-time reinforcement learning problem. We model the task as a Markov decision process
(MDP) $\mathcal{M_{\mathrm{MDP}}} = \langle \mathcal{S}, \mathcal{A}, \mathcal{T}, \mathcal{R}, \gamma \rangle$,
where $\mathcal{S}$ is the state space, $\mathcal{A}$ the action space,
$\mathcal{T}(\bm{s}_{t+1}\mid\bm{s}_t,\bm{a}_t)$ the transition dynamics induced
jointly by the robot dynamics and the obstacle motion,
$\mathcal{R}:\mathcal{S}\times\mathcal{A}\to\mathbb{R}$ the reward function, and
$\gamma\in[0,1)$ the discount factor. The objective is to learn a policy
$\pi_\theta(\bm{a}_t\mid\bm{o}_t)$ that maximizes the expected discounted return
$J(\theta) = \mathbb{E}_{\pi_\theta}[\sum_{t=0}^{\infty}\gamma^t r_t]$. In this work, the policy's objective is to avoid dynamic obstacles while maintaining stability without failure. A failure is defined as any non-foot body part of the humanoid making contact with the ground.

\textbf{Observation space.}
At each control step $t$, the policy receives an observation
$\bm{o}_t = [\bm{s}_t^{p},\,\bm{s}_t^{o}]$, comprising
proprioceptive and obstacle states. The proprioceptive
state $\bm{s}_t^{p} = [\bm{\omega}_t,\phi_t,\theta_t,\bm{q}_t,\dot{\bm{q}}_t,\bm{a}_{t-1}]$
contains the base angular velocity $\bm{\omega}_t\in\mathbb{R}^3$, the base roll
$\phi_t$ and pitch $\theta_t$, the joint positions $\bm{q}_t\in\mathbb{R}^N$ and
velocities $\dot{\bm{q}}_t\in\mathbb{R}^N$, and the previous action
$\bm{a}_{t-1}\in\mathbb{R}^N$, where $N$ is the number of actuated joints. The obstacle state
$\bm{s}_t^{o} = [x_t^{o},y_t^{o},z_t^{o}]\in\mathbb{R}^3$ gives the obstacle's 3D position in the robot base frame.

\textbf{Action space.}
The policy outputs an action $\bm{a}_t\in\mathbb{R}^N$ specifying joint position
offsets relative to a nominal standing configuration. The desired joint targets
are $\bm{q}_t^{\mathrm{target}} = \alpha\,\bm{a}_t + \bm{q}^{\mathrm{default}}$,
where $\alpha$ is the action scaling factor and $\bm{q}^{\mathrm{default}}$ the
default standing configuration. These targets are tracked by joint-level
proportional--derivative (PD) controllers producing torques
$\bm{\tau}_t = K_p(\bm{q}_t^{\mathrm{target}} - \bm{q}_t) - K_d\dot{\bm{q}}_t$,
with stiffness and damping gains $K_p$ and $K_d$. 

\subsection{Threat-aware avoidance policy training}
We approximate the dynamic obstacle as a sphere with center position $\bm{p}_t^{o}\in\mathbb{R}^3$,
velocity $\bm{v}_t^{o}\in\mathbb{R}^3$, and radius $R^o$. The humanoid is represented by
$K=10$ key bodies $\mathcal{B}=\{1,\ldots,K\}$ (the head, torso, and upper and
lower links of both arms and legs). Each body $i$ is represented by a sphere with
center position $\bm{p}_t^{i}\in\mathbb{R}^3$, velocity $\bm{v}_t^{i}\in\mathbb{R}^3$, radius $R^i$,
and safety margin $m = 0.5$~$\mathrm{m}$. This representation permits efficient per-body threat
estimation using sphere--sphere geometry. Let $R_{\mathrm{safe}}^i=R^o+R^i+m$ be the radius of the safety region.

\textbf{Predictive threat.}
We first estimate whether the obstacle will enter the safety region of each
body within a prediction horizon $H = 3$~$\mathrm{s}$. Under a constant-velocity approximation, we compute the time of closest approach and its
validity mask as
\begin{equation}
\tau_t^{i*}=-\frac{(\Delta\bm{p}_t^{i})^\top\bm{w}_t^{i}}
{\lVert\bm{w}_t^{i}\rVert^2}, \qquad
M_t^i=\mathbb{I}(0<\tau_t^{i*}<H),
\label{eq:closest_approach_time}
\end{equation}
where $\Delta\bm{p}_t^{i}=\bm{p}_t^{o}-\bm{p}_t^{i}$ and $\bm{w}_t^{i}=\bm{v}_t^{o}-\bm{v}_t^{i}$ are the relative position and velocity from body $i$ to the obstacle, respectively. In this formulation, obstacles that are already receding or whose closest approach lies
beyond $H$ do not activate the predictive threat. The additional displacement required
to keep body $i$ outside the region is
\begin{equation}
\delta_t^i=M_t^i\max(0,R_i^{\mathrm{safe}}-d_t^{i*}),
\label{eq:required_displacement}
\end{equation}
where $d_t^{i*} = \lVert\Delta\bm{p}_t^{i}+\tau_t^{i*}\bm{w}_t^{i}\rVert$ is the distance of closest approach. The required acceleration magnitude and the resulting predictive threat are computed as
\begin{equation}
\begin{aligned}
a_t^{\mathrm{req},i}
&=\frac{2\delta_t^i}
{{\tau_t^{i*}}^2},\qquad
S_t^i
&=1-\exp\!\left[-\left(
\frac{a_t^{\mathrm{req},i}}{a_{\mathrm{ref}}}\right)^2\right],
\end{aligned}
\label{eq:predictive_threat}
\end{equation}
where $a_{\mathrm{ref}} = 2$~$\mathrm{m/s^2}$ is a reference acceleration. Consequently, $S_t^i\in[0,1]$ jointly captures the
future collision geometry, remaining reaction time, and required evasive
effort. 

\textbf{Near-field threat.}
The predictive threat is complemented by a bounded near-field threat that depends
only on the current surface clearance $c_t^i=\lVert\Delta\bm{p}_t^i\rVert-R_{\mathrm{safe}}^i$. We map this clearance to
a bounded near-field threat using a one-sided Gaussian kernel,
\begin{equation}
D_t^{i}
=\exp\!\left[-\left(
\frac{\max(c_t^i,0)}{\sigma_{\mathrm{near}}}\right)^2\right],
\label{eq:near_field_threat}
\end{equation}
where $\sigma_{\mathrm{near}} = 0.5$~$\mathrm{m}$ controls the spatial decay. The threat equals
one when the obstacle enters the safety region and smoothly decreases as the obstacle moves
away. 

\textbf{Threat-aware avoidance reward.}
To encourage the humanoid to dodge dynamic obstacles, we design rewards based on the predictive and near-field threats. We define the body escape reward as
\begin{equation}
r_t^{\mathrm{escape}}
=
\sum_{i\in\mathcal{B}}
\left[
S_t^i
\tanh\!\left(
\frac{
\max(0,\bm{v}_t^i\!\cdot\bm{e}_t^{\mathrm{pred},i})
}{
1.5
}
\right)
\right],
\label{eq:reward_escape}
\end{equation}
where
\begin{equation}
\bm{e}_t^{\mathrm{pred},i}=-\frac{\Delta\bm{p}_t^{i}+\tau_t^{i*}\bm{w}_t^{i}}
{\lVert\Delta\bm{p}_t^{i}+\tau_t^{i*}\bm{w}_t^{i}\rVert},\qquad
\end{equation}
is the desired body escape direction, which points away from the predicted closest-approach point. The reward $r_t^{\mathrm{escape}}$ encourages threatened bodies to move away from the predicted obstacle trajectory, with each body's contribution weighted by $S_t^i$.
We also penalize obstacle proximity to the robot's bodies using the near-field threat measure, with a larger penalty for obstacles within the safety region. The body distance reward is calculated as
\begin{equation}
r_t^{\mathrm{distance}}
=
\sum_{i\in\mathcal{B}}
\begin{cases}
-D_t^i, & c_t^i \ge 0, \\[2mm]
-\left(1+\dfrac{|c_t^i|}{\sigma_{\mathrm{near}}}\right),
& c_t^i < 0.
\end{cases}
\label{eq:keybody_distance_reward}
\end{equation}

While the body escape and distance rewards encourage dodging behavior, they do not explicitly guide the robot's overall movement.
Therefore, we further define a root-body escape-direction reward,
\begin{equation}
r_t^{\mathrm{root\text{-}dir}} = \mathbb{I}(S_t^{\mathrm{root}}>0)
\phi\!\left(
\hat{\bm{v}}_t^{\mathrm{root}}\!\cdot
\bm{e}_{t,xy}^{\mathrm{pred,root}}\right),
\label{eq:root_direction_reward}
\end{equation}
where $\hat{\bm{v}}_t^{\mathrm{root}} =
\bm{v}_{t,xy}^{\mathrm{root}}/
\lVert\bm{v}_{t,xy}^{\mathrm{root}}\rVert$ is the root velocity direction in
the $xy$ plane. The asymmetric
direction-scoring function is
\begin{equation}
\phi(x)=\max(x,0)-\lambda_{\mathrm{wrong}}\max(-x,0),
\label{eq:direction_score}
\end{equation}
where $\lambda_{\mathrm{wrong}}=5$ penalizes root motion in the
wrong direction. Regularization rewards are included to help the robot maintain a stable posture and produce smooth joint-level actions. When the overall threat $T = \max_{i\in\mathcal{B}}[1-(1-S_t^i)(1-D_t^i)]$ is zero (in practice, we use $T<10^{-3}$ to disregard small nonzero tails in $D_t^i$), we apply recovery rewards that help the robot return to a stable standing posture. We further include a capture-point reward to encourage stable recovery after evasive maneuvers:
\begin{equation}
r^{\mathrm{capture}}
=
\exp\left(
-\left\|
\bm{p}^{\mathrm{capture}}
-
\bm{p}^{\mathrm{supp}}
\right\|_2^2/\sigma_{\mathrm{cp}}^2
\right),
\label{eq:capture_point_reward}
\end{equation}
where $\bm{p}^{\mathrm{capture}}=\bm{p}_{xy}^{\mathrm{root}}+\bm{v}_{xy}^{\mathrm{root}}/\omega_0$ is the instantaneous capture point, $\omega_0=\sqrt{g/z^{\mathrm{root}}}$ is the natural frequency of the linear inverted pendulum model, $g$ is the gravitational acceleration, and $z^{\mathrm{root}}$ is the robot root height. The support position $\bm{p}^{\mathrm{supp}}=(\bm{p}_{xy}^{L}+\bm{p}_{xy}^{R})/2$ is approximated by the midpoint of the left and right feet, and $\sigma_{\mathrm{cp}} = 0.12$~$\mathrm{m}$ controls the sensitivity of the reward to the capture-point tracking error.

The detailed reward terms and corresponding weights are shown in Table~\ref{tab:reward_terms1}.

\begin{table}
\vspace{5pt}
\centering
\caption{Reward functions for policy training and their corresponding weights.}
\renewcommand{\arraystretch}{1.5}
\resizebox{0.99\linewidth}{!}{
\begin{tabular}{l c c}
\hline
\textbf{Name} & \textbf{Expression} & \textbf{Weight} \\
\hline
\multicolumn{3}{l}{\textbf{Dodge rewards ($T>0$)}}\\
Body escape & $
r^{\mathrm{escape}} $ (Eq.~\ref{eq:reward_escape})& 40.0\\
Body distance & $
r^{\mathrm{distance}} $ (Eq.~\ref{eq:keybody_distance_reward})& 50.0\\
Root-body escape direction &$
r^{\mathrm{root-dir}} $ (Eq.~\ref{eq:root_direction_reward})& 60.0\\
\hline
\multicolumn{3}{l}{\textbf{Recovery rewards in the absence of threats ($T=0$)}}\\
Root velocity&$\exp\left(
-\left\|\bm{v}_{xy}^{\mathrm{root}}\right\|_2^2 / 0.5^2
\right)$& 10.0\\
Torso orientation&$
\left(\phi^{\mathrm{torso}}\right)^2
+
\left(\theta^{\mathrm{torso}}\right)^2$&-20.0\\
Joint deviation&$\sum_{i=1}^{N}
\left(
q_i-q_{i, \mathrm{default}}
\right)^2$& -40.0\\
Capture point&$r^{\mathrm{capture}}$ (Eq.~\ref{eq:capture_point_reward})& 10.0\\
\hline
\multicolumn{3}{l}{\textbf{Regularization rewards}}\\
AMP reward&$r^{\mathrm{AMP}}$& 10.0\\
Angular velocity penalty&$\left\lVert \omega_{xy}^{\mathrm{root}} \right\rVert^2$& -0.2\\
Joint velocities&$\|\boldsymbol{\dot{\bm{q}}}\|_2^2$& -0.001\\
Joint accelerations&$\left\lVert (\dot{\bm{q}}_{t} - \bm{\dot{q}}_{t-1})/dt \right\rVert^2$&-2.5e-7\\
Action rate&$\left\lVert ({\bm{a}}_{t} - {\bm{a}}_{t-1})/dt \right\rVert^2$&-0.1\\
Action acceleration&$\left\|
\bm{a}_t - 2\bm{a}_{t-1} + \bm{a}_{t-2} \right\|_2^2$&-0.05\\
Joint position limit&$
\sum_{i=1}^{N} \!\max\!\left( \lvert q_i \rvert - 0.9 \cdot q_{i,\mathrm{lim}},\,0 \right)
$& -10.0\\
Joint velocity limit&$
\sum_{i=1}^{N} \!\max\!\left( \lvert \dot{q}_i \rvert - 0.9 \cdot \dot{q}_{i,\mathrm{lim}},\,0 \right)
$& -5.0\\
Joint torque limit&$
\sum_{i=1}^{N} \!\max\!\left( \lvert \tau_i \rvert - 0.85 \cdot \tau_{i,\mathrm{lim}},\,0 \right)
$& -5.0\\
Foot slip&$c_{\mathrm{L}}
\left\| \bm{v}_{xy}^{\mathrm{L}}
\right\|_2^2 + c_{\mathrm{R}} \left\|
\bm{v}_{xy}^{\mathrm{R}}
\right\|_2^2$& -10.0\\
Foot clearance&$\sum_{i\in\{\mathrm{L},\mathrm{R}\}}
(1-c_i)\,
\operatorname{clip}\!\left(h_i,0,0.25\right)$& 20.0\\
\hline
\end{tabular}}
\label{tab:reward_terms1}
\vspace{2pt}
\begin{minipage}{0.99\linewidth}
\footnotesize
$c_{\mathrm{L}}$ and $c_{\mathrm{R}}$ denote the left and right foot contact states, respectively, and
$\bm{v}^{\mathrm{L}}$ and $\bm{v}^{\mathrm{R}}$ denote the foot velocities. $h_i$ denotes the height of foot $i$ above the ground.
\end{minipage}
\vspace{-10pt}
\end{table}

\subsection{Adversarial motion priors}
We use adversarial motion priors (AMP)~\cite{peng2021amp} to regularize the learned dodging behaviors toward dynamically feasible and natural human-like motions. We use 90 human evasive motion clips from the dataset in~\cite{bones2026seed}, spanning four dodge directions: left, right, backward, and forward. For each direction, multiple motion styles are included to capture diverse realizations of the evasive behavior, such as dodging with an upright torso or combining the dodge with downward motion of the head and torso. All motion clips are retargeted to the robot's morphology using the method in~\cite{joao2025gmr}.

\textbf{Discriminator training.}
We employ four direction-specific AMP discriminators, each associated with one cardinal dodge direction: left, right, forward, or backward. Each discriminator is trained using joint-state transitions from the reference motion dataset corresponding to its associated dodge direction and transitions generated by the policy. At each timestep, the discriminator is selected according to the desired root escape direction $\bm{e}_{t,xy}^{\mathrm{pred,root}}$. If $\bm{e}_{t,xy}^{\mathrm{pred,root}}=\bm{0}$, one of the four discriminators is selected uniformly at random. Otherwise, $\bm{e}_{t,xy}^{\mathrm{pred,root}}$ is mapped to the nearest cardinal direction, and the corresponding discriminator is activated. For discriminator $D_k$, where $k \in \left\{ \mathrm{left}, \mathrm{right}, \mathrm{forward}, \mathrm{backward} \right\}$, let $\bm{x}_t=(\bm{q}_t,\bm{q}_{t+1})$ denote a joint-state transition. The discriminator is trained using the least-squares adversarial objective
\begin{equation}
\begin{aligned}
\min_{D_k}\quad
&\mathbb{E}_{\bm{x}_t\sim d_M^k}
\left[\left(D_k(\bm{x}_t)-1\right)^2\right]\\
&+\mathbb{E}_{\bm{x}_t\sim d_{\pi}^k}
\left[\left(D_k(\bm{x}_t)+1\right)^2\right],
\end{aligned}
\label{eq:multi_amp}
\end{equation}
where $d_M^k$ denotes the reference transition distribution for dodge direction $k$, and $d_{\pi}^k$ denotes the policy transition distribution assigned to discriminator $D_k$ according to the selected root escape direction. The policy receives the AMP reward from the selected discriminator $D_k$ as $r_t^{\mathrm{AMP}} =
\max \left[0,\,1 - 0.25 \left(D_k(\bm{q}_t,\bm{q}_{t+1})-1 \right)^2 \right]$.

\subsection{Training details}
\textbf{Training environment setup.}
We train the policies in the mjlab simulator \cite{zakka2026mjlab} using 4096 parallel environments and evaluate our framework on two humanoid platforms, Unitree G1 and EngineAI PM01. To expose the policies to diverse dynamic obstacles, an obstacle is initialized at a randomly sampled position relative to the robot. Its initial velocity is generated using one of two schemes: 1) the obstacle is directed toward a randomly selected body part with a randomly sampled speed; or 2) the obstacle is aimed toward the robot root in the horizontal plane while an additional vertical velocity component is applied, producing a parabolic projectile trajectory. We further randomize the obstacle radius and the gravitational acceleration acting on the obstacle to increase the diversity of projectile dynamics. The detailed domain randomization settings are summarized in Table~\ref{tab:domain_rand}.

\textbf{Policy network.}
The policy actor and critic are parameterized by multilayer perceptrons (MLPs) with hidden-layer sizes $[512,256,128]$. Both networks receive the current observation $\bm{o}_t$ together with a 20-step observation history. The history is processed by a temporal encoder consisting of a per-frame linear embedding, followed by a two-layer 1-D temporal convolutional network and a final linear projection that produces a 128-dimensional latent representation. Notably, the obstacle velocity is not provided to the actor; instead, projectile motion is inferred implicitly from the history of obstacle positions. In addition, the critic receives the privileged observation $\bm{o}'_t=[\bm{D}_t,\bm{S}_t,c_{\mathrm{L}},c_{\mathrm{R}},\bm{v}_t,\bm{v}_t^{o},\bm{p}_t^{b},r^{o}]$, where $\bm{D}_t$ and $\bm{S}_t$ denote the near-field and predictive threat values, respectively, for all key bodies; $c_{\mathrm{L}}$ and $c_{\mathrm{R}}$ indicate the left and right foot contact states; $\bm{v}_t$ is the robot root velocity; $\bm{v}_t^{o}$ is the obstacle velocity; $\bm{p}_t^{b}$ contains the positions of the robot's key bodies expressed in the robot base frame; and $r^{o}$ denotes the obstacle radius. We use the proximal policy optimization (PPO) algorithm for policy optimization~\cite{schulman2017ppo}. The policy is trained for 30,000 iterations.

\begin{table}[!t]
\vspace{5pt}
\centering
\caption{Domain randomization parameters. All quantities are sampled uniformly within the specified ranges.}
\begin{tabular}{l l l}
\hline
\textbf{Category} & \textbf{Term} & \textbf{Range} \\
\hline

\multirow{4}{*}{\textbf{Dynamics}}
    & Friction coefficient
    & $[0.1,\,2.0]$ \\
    & Joint $K_p,K_d$ scaling
    & $[0.8,\,1.2]\times$ \\
    & Joint position bias
    & $[-0.05,\,0.05]~\mathrm{rad}$ \\
    & Torso COM offset
    & $[-0.05,\,0.05]~\mathrm{m}$ \\
\hline

\multirow{6}{*}{\textbf{Sensor noise}}
    & Base angular velocity $\bm{\omega}$
    & $[-0.2,\,0.2]~\mathrm{rad/s}$ \\
    & Base roll/pitch $(\phi,\theta)$
    & $[-0.25,\,0.25]~\mathrm{rad}$ \\
    & Joint positions $\bm{q}$
    & $[-0.02,\,0.02]~\mathrm{rad}$ \\
    & Joint velocities $\dot{\bm{q}}$
    & $[-1.5,\,1.5]~\mathrm{rad/s}$ \\
    & Previous action $\bm{a}_{t-1}$
    & $[-0.2,\,0.2]$ \\
    & Obstacle position $\bm{s}^{o}$
    & $[-0.2,\,0.2]~\mathrm{m}$ \\
\hline

\multirow{3}{*}{\textbf{Disturbances}}
    & Linear velocity push $\bm{v}_{xy}^{\mathrm{push}}$
    & $[-1.0,\,1.0]~\mathrm{m/s}$ \\
    & Roll/pitch velocity push $\bm{\omega}_{\phi\theta}^{\mathrm{push}}$
    & $[-0.6,\,0.6]~\mathrm{rad/s}$ \\
    & Yaw velocity push $\omega_{\psi}^{\mathrm{push}}$
    & $[-0.8,\,0.8]~\mathrm{rad/s}$ \\
\hline

\multirow{6}{*}{\shortstack{\textbf{Obstacle}\\\textbf{initial state}}}
    & Horizontal position $(x^{o},y^{o})$
    & $[-6,\,6]~\mathrm{m}$ \\
    & Height $z^{o}$
    & $[0,\,1.8]~\mathrm{m}$ \\
    & Launch speed $\|\bm{v}_{xy}^{o}\|$
    & $[2,\,16]~\mathrm{m/s}$ \\
    & Launch speed $\|\bm{v}_{z}^{o}\|$
    & $[1,\,6]~\mathrm{m/s}$ \\
    & Obstacle radius $R^{o}$
    & $[0.05,\,0.125]~\mathrm{m}$ \\
    & Gravity scale 
    & $[0.0,\,1.0]g$ \\
\hline
\end{tabular}
\label{tab:domain_rand}
\vspace{-10pt}
\end{table}

\section{Simulation experiments}

In this section, we present simulation results, analyze our method, and compare it with baseline methods. Specifically, we answer the following questions: \textbf{Q1} Does our method achieve better avoidance performance than the baselines across a wide range of dynamic obstacle states? \textbf{Q2} Does our policy exhibit more diverse avoidance motions, and does this diversity lead to better obstacle avoidance performance? \textbf{Q3} How do our reward design and the use of multiple AMP discriminators affect performance?

\subsection{Metrics and baselines}

We consider the following performance metrics:

\textbf{1) Success rate $(\%)$}: $
SR =
\frac{N_{\mathrm{success}}}{N_{\mathrm{obstacle}}}
\times 100\%,
$
where $N_{\mathrm{success}}$ is the number of times the robot dodges the obstacle and maintains stability without falling, and $N_{\mathrm{obstacle}}$ is the total number of dynamic obstacle occurrences.

\textbf{2) Recovery time} $(\mathrm{s})$:
$T_{\mathrm{recovery}} = t_{\mathrm{stable}}-t_{\mathrm{onset}}$,
where $t_{\mathrm{onset}}$ denotes the timestep at which the robot initiates an evasive maneuver, identified when the root velocity satisfies $\|\bm{v}^{\mathrm{root}}\|>0.15~\mathrm{m/s}$. The timestep $t_{\mathrm{stable}}$ is defined as the beginning of the first continuous $1$-s interval after the threat becomes zero during which $\|\bm{v}^{\mathrm{root}}\|<0.15~\mathrm{m/s}$, $|\theta^{\mathrm{torso}}|<0.2~\mathrm{rad}$, and $|\phi^{\mathrm{torso}}|<0.2~\mathrm{rad}$ are all satisfied.

\textbf{3) Average power} $(\mathrm{W})$: The average motor power consumed by all actuated joints during a successful evasive maneuver is computed as
\[
P_{\mathrm{motor}}
=
\frac{1}{t_{\mathrm{stable}}-t_{\mathrm{onset}}}
\int_{t_{\mathrm{onset}}}^{t_{\mathrm{stable}}}
\sum_i
\left|
\tau_i(t)\dot{q}_i(t)
\right|\,dt.
\]

We compare our approach against two whole-body dynamic obstacle avoidance baselines:

\textbf{1) REBot}~\cite{xu2025rebot}:  
A reinforcement learning framework for reflexive dynamic obstacle avoidance in quadruped robots using curriculum learning and policy switching. We reimplement its training pipeline for humanoid robots.

\textbf{2) PAC-MAN}~\cite{yang2026pacman}:  
Developed concurrently with our work, it presents a perception-aware reinforcement learning framework for humanoid dynamic obstacle avoidance that combines CBF-inspired reward shaping with AMP-based motion regularization. For a fair comparison, we retrain PAC-MAN using our human motion dataset and use the state-oracle perception option in which the ground-truth obstacle states are observable to the actor.

For a fair comparison, both baselines are trained using the same obstacle initialization settings as our method.

\begin{figure}[!t]
\vspace{8pt}
\centering
\includegraphics[width=1\linewidth]{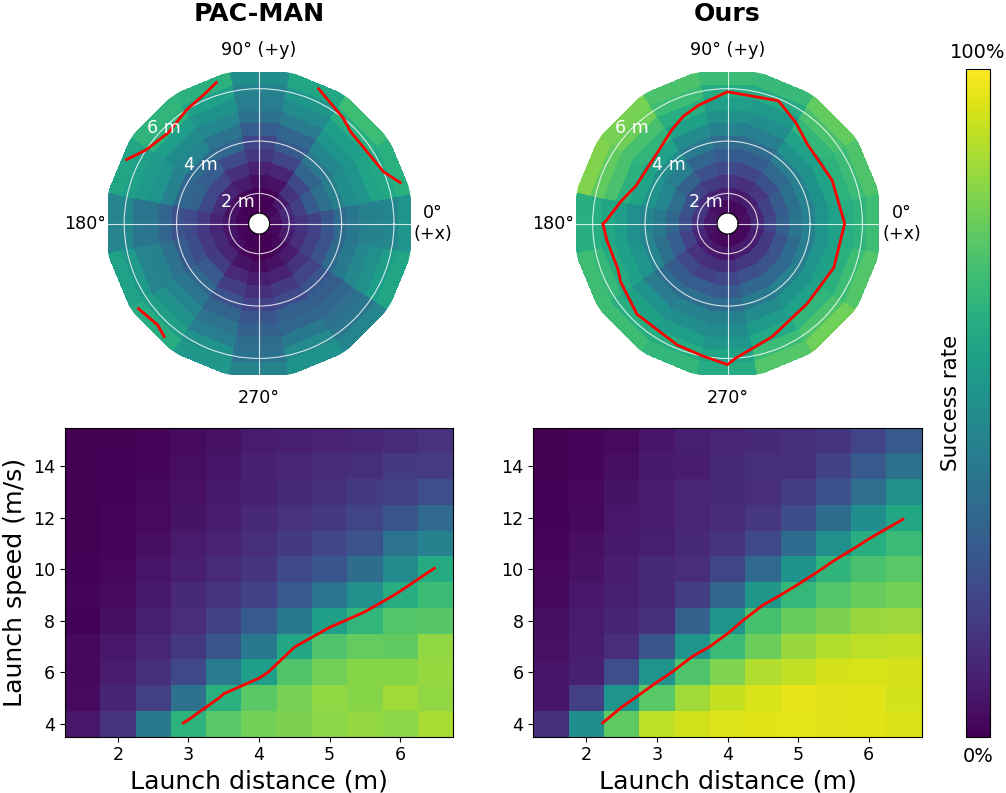}
\caption{Top: Polar heatmaps of obstacle avoidance success rates across different launch distances and approach directions relative to the robot's heading. Bottom: Success-rate heatmaps across different obstacle launch speeds and distances. The red contour indicates the 60\% success-rate boundary.}
\label{fig:heatmap}
\vspace{-1.0em}  
\end{figure}

\begin{table*}[t]
\vspace{5pt}
\centering
\caption{Dodge performance under different time-to-hit ($\mathrm{TTH}$) ranges in simulation.
Recovery time and average motor power are reported as mean~$\pm$~standard deviation. Bold indicates the best result for each robot and $\mathrm{TTH}$ range.}
\begin{tabular}{c l|ccc|ccc}
\toprule
&
&
\multicolumn{3}{c|}{\textbf{Unitree G1}}
&
\multicolumn{3}{c}{\textbf{EngineAI PM01}}
\\
\cmidrule(lr){3-5}
\cmidrule(lr){6-8}

\textbf{\boldmath$\mathrm{TTH}$ (s)}
& \textbf{Method}
& $\bm{SR}$ $(\%)$
& $\bm{T_{\mathrm{recovery}}}$ $(\mathrm{s})$
& $\bm{{P}_{\mathrm{motor}}}$ $(\mathrm{W})$
& $\bm{SR}$ $(\%)$
& $\bm{T_{\mathrm{recovery}}}$ $(\mathrm{s})$
& $\bm{{P}_{\mathrm{motor}}}$ $(\mathrm{W})$
\\
\midrule

\multirow{3}{*}{$<0.4$}
& Ours
& \textbf{11.8} & $1.194 \pm 0.356$ & $569.6 \pm 124.5$
& \textbf{14.2} & $\bm{1.197 \pm 0.439}$ & $586.3 \pm 145.9$ \\
& REBot~\cite{xu2025rebot}
& 4.8 & $1.772 \pm 0.468$ & $645.3 \pm 93.5$
& 8.1 & $2.293 \pm 0.972$ & $830.9 \pm 266.6$ \\
& PAC-MAN~\cite{yang2026pacman}
& 5.9 & $\bm{0.825 \pm 0.467}$ & $\bm{316.7 \pm 77.2}$
& 3.9 & $1.788 \pm 1.382$ & $\bm{335.9 \pm 139.7}$ \\
\midrule

\multirow{3}{*}{$0.4$--$0.8$}
& Ours
& \textbf{64.6} & $1.438 \pm 0.900$ & $587.7 \pm 105.7$
& \textbf{55.5} & $1.369 \pm 0.666$ & $590.0 \pm 129.3$ \\
& REBot~\cite{xu2025rebot}
& 10.0 & $1.735 \pm 0.443$ & $618.7 \pm 86.1$
& 9.8 & $2.162 \pm 0.575$ & $834.1 \pm 228.0$ \\
& PAC-MAN~\cite{yang2026pacman}
& 36.8 & $\bm{0.570 \pm 0.337}$ & $\bm{319.8 \pm 61.0}$
& 33.6 & $\bm{1.097 \pm 0.939}$ & $\bm{396.8 \pm 104.8}$ \\
\midrule

\multirow{3}{*}{$0.8$--$1.2$}
& Ours
& \textbf{95.1} & $1.119 \pm 0.300$ & $507.9 \pm 95.8$
& \textbf{94.2} & $1.298 \pm 0.343$ & $580.3 \pm 117.2$ \\
& REBot~\cite{xu2025rebot}
& 34.0 & $1.826 \pm 0.396$ & $553.0 \pm 67.6$
& 17.3 & $2.416 \pm 0.813$ & $777.3 \pm 182.9$ \\
& PAC-MAN~\cite{yang2026pacman}
& 79.2 & $\bm{0.244 \pm 0.210}$ & $\bm{312.8 \pm 58.5}$
& 79.5 & $\bm{0.589 \pm 0.319}$ & $\bm{395.3 \pm 76.8}$ \\
\midrule

\multirow{3}{*}{$\geq 1.2$}
& Ours
& \textbf{95.6} & $1.190 \pm 0.363$ & $399.3 \pm 71.2$
& \textbf{97.2} & $1.303 \pm 0.341$ & $486.0 \pm 104.8$ \\
& REBot~\cite{xu2025rebot}
& 53.1 & $2.008 \pm 0.565$ & $522.4 \pm 102.7$
& 43.9 & $2.190 \pm 0.825$ & $752.8 \pm 174.4$ \\
& PAC-MAN~\cite{yang2026pacman}
& 81.9 & $\bm{0.262 \pm 0.192}$ & $\bm{305.2 \pm 58.7}$
& 86.6 & $\bm{0.237 \pm 0.150}$ & $\bm{386.0 \pm 73.7}$ \\

\bottomrule
\end{tabular}
\label{tab:performance_tth}
\end{table*}

\subsection{Evaluation of avoidance performance}
To evaluate avoidance performance across a broad range of obstacle states, we systematically vary the obstacle launch speed $v^o \in [4,5,\ldots,15]$ m/s, the approach direction relative to the robot heading $\theta^o \in [0,\tfrac{\pi}{8},\ldots,\tfrac{15\pi}{8}]$ rad, and the horizontal launch distance $d^o \in [1.0,1.5,\ldots,6.5]$ m. This results in $12 \times 16 \times 12 = 2304$ distinct obstacle configurations. Each configuration is evaluated over 100 trials. In every trial, the targeted body part is sampled uniformly from the robot's key bodies, while the obstacle gravity scale is independently randomized within $[0,1]g$, covering trajectories from straight-line motion to fully ballistic flight.

We compute the success rate ($SR$) for each obstacle configuration. Fig.~\ref{fig:heatmap} (top) reports $SR$ averaged over launch speeds $v^o$ for each launch distance-direction pair $(d^o,\theta^o)$. PAC-MAN exhibits pronounced directional sensitivity, with noticeable performance degradation for several frontal, rear, and lateral approach directions. In contrast, our method maintains a consistently higher success rate over nearly the entire angular range, indicating stronger robustness to obstacle approach direction and more uniform whole-body evasive capability.
Fig.~\ref{fig:heatmap} (bottom) reports $SR$ averaged over approach directions $\theta^o$ for each launch speed-distance pair $(v^o,d^o)$. As expected, both methods become less successful as the obstacle speed increases or the launch distance decreases, reducing the available reaction time. However, our method consistently achieves higher success rates across the speed-distance space. The red contour denotes the $60\%$ success-rate boundary. Compared with PAC-MAN, our method shifts this boundary toward substantially more challenging obstacle configurations, demonstrating improved tolerance to faster projectiles and shorter reaction distances.

We group the obstacle launch states into four nominal time-to-hit ($\mathrm{TTH}$) categories, where $\mathrm{TTH}=d^o/v^o$, and summarize the results in Table~\ref{tab:performance_tth}. Our method consistently achieves the highest success rate ($SR$) across all $\mathrm{TTH}$ ranges and on both humanoid platforms. The advantage becomes particularly pronounced as more reaction time becomes available. This indicates that the learned policy effectively exploits additional reaction time and generalizes consistently across two humanoid platforms. REBot performs substantially worse, reaching only around $50\%$ for $\mathrm{TTH}\geq1.2$~s, while its success rate drops rapidly as $\mathrm{TTH}$ decreases. REBot also generally exhibits longer recovery times and considerably higher motor power consumption, particularly on PM01, suggesting that its reflexive evasion training pipeline, originally developed for quadrupeds without AMP-based motion regularization, transfers poorly to humanoid whole-body avoidance. PAC-MAN performs significantly better than REBot and has the lowest average mechanical power and generally the shortest recovery time. However, its avoidance success remains below ours in every $\mathrm{TTH}$ range. Overall, our method provides the strongest obstacle avoidance performance at the cost of higher motor power consumption and slower post-evasion recovery than PAC-MAN.

\begin{figure}
    \centering
    \includegraphics[width=1.0\linewidth]{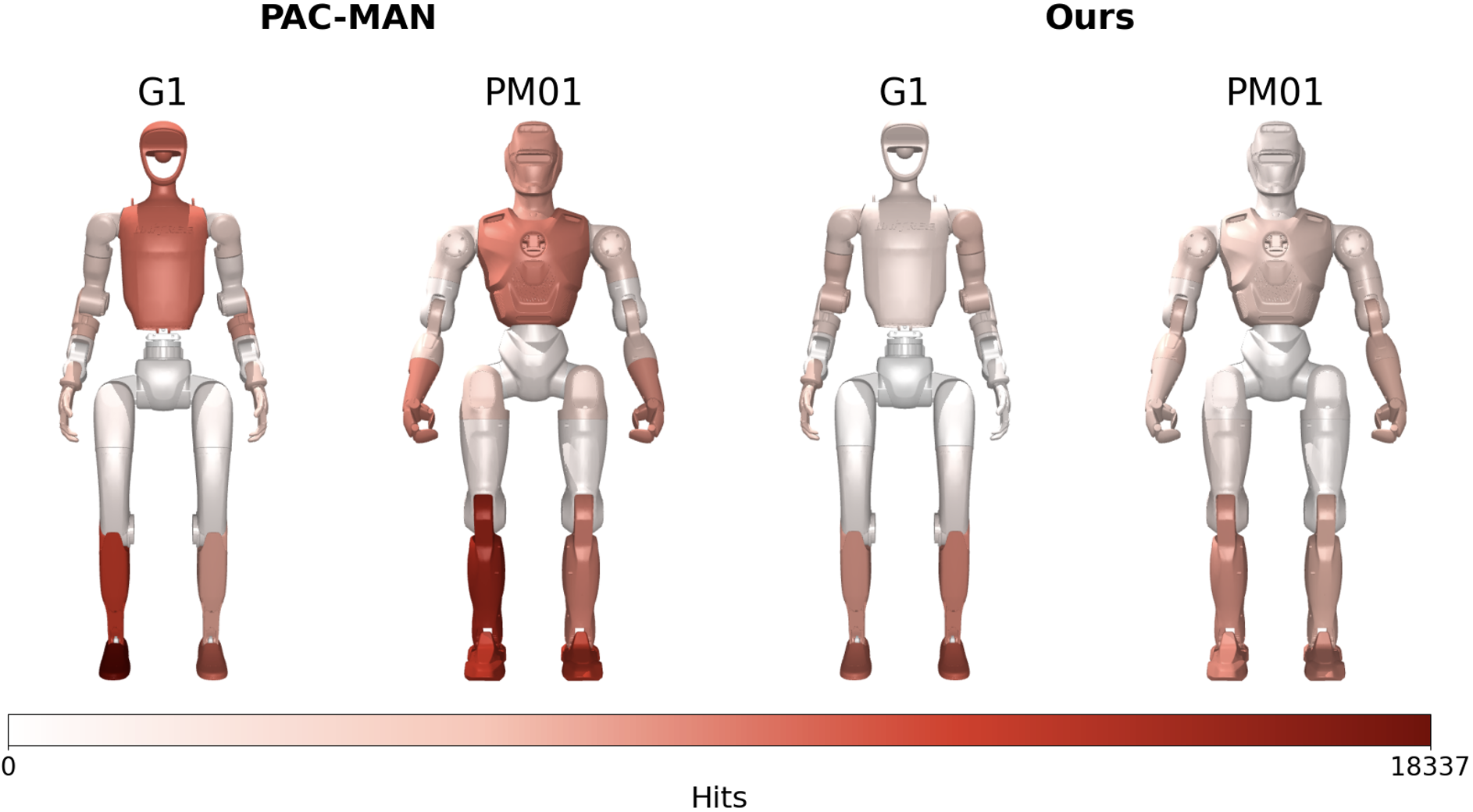}
    \caption{Visualization of obstacle collision frequency on the G1 and PM01 humanoids during the simulation evaluation.}
    \label{fig:hitmap}
    \vspace{-1.0em}  
\end{figure}

Fig.~\ref{fig:hitmap} visualizes the collision frequency of individual body parts for the G1 and PM01 humanoids during evaluation. For PAC-MAN, collisions are concentrated primarily around the head, torso, and lower limbs, while the pelvis and upper legs experience comparatively fewer contacts. In contrast, our method substantially reduces collision frequency across most upper-body regions, with the remaining contacts occurring mainly around the lower legs and feet. This concentration is likely due to the functional constraints of the lower limbs during dynamic evasion: the feet must prioritize support and balance maintenance, leaving less freedom for aggressive collision-avoidance motions than the upper body.

\subsection{Analysis of avoidance motions}
\begin{figure}
    \vspace{8pt}
    \centering
    \includegraphics[width=1.0\linewidth]{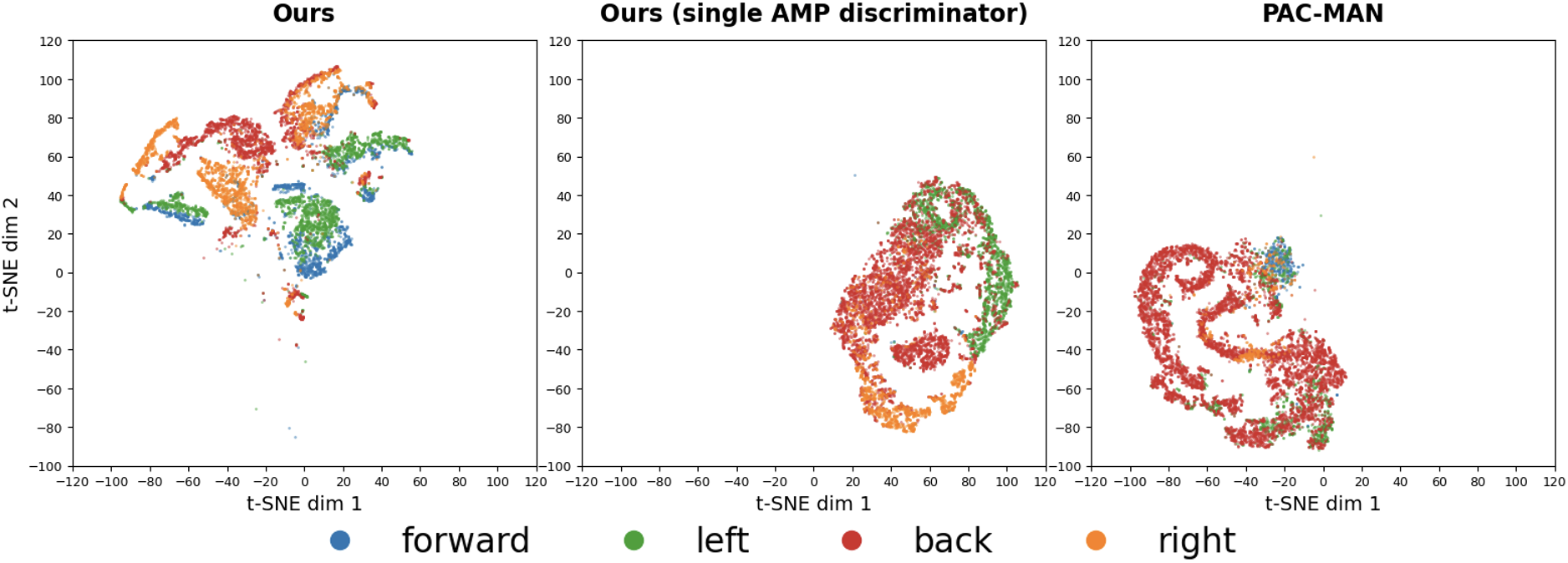}
    \caption{t-SNE visualization of avoidance motions from the evaluation rollouts. }
    \label{fig:tsne}
    \vspace{-1.0em}  
\end{figure}

To characterize the diversity of learned evasive behaviors, we collect motion windows generated during evaluation across a wide range of obstacle configurations. Each window contains the 20 most recent frames, where each frame is represented by $\bm{f}_t=[\bm{q}_t,\bm{v}_t^{\mathrm{root},h},\bm{\omega}_t^{\mathrm{root}},z_t^{\mathrm{root}}]$. Each motion window is labeled according to the robot's movement direction during the window. Fig.~\ref{fig:tsne} shows the t-SNE embeddings of 10000 motion windows randomly sampled during evaluation for our method, our method with a single AMP discriminator, and PAC-MAN. Our method exhibits the broadest and most balanced distribution across forward, backward, leftward, and rightward evasive motions, indicating greater behavioral diversity. In contrast, PAC-MAN is strongly dominated by backward dodging, with comparatively few forward and lateral motions. Using only a single AMP discriminator also leads to clear mode imbalance, particularly with very limited forward-dodge behaviors. Introducing direction-specific AMP discriminators produces more distinct and better represented motion modes across all four dodge directions, demonstrating their effectiveness in promoting diverse evasive behaviors. 

\subsection{Ablation studies}
To assess the contributions of individual reward terms and direction-specific motion priors, we compare the full framework with the following five ablated variants: 
\textbf{1) w/o body escape reward} $r^{\mathrm{escape}}$, 
\textbf{2) w/o body distance reward} $r^{\mathrm{distance}}$, 
\textbf{3) w/o root-direction reward} $r^{\mathrm{root-dir}}$, 
\textbf{4) w/o capture-point reward} $r^{\mathrm{capture}}$, and 
\textbf{5) single AMP discriminator}, which replaces the four direction-specific discriminators with one shared discriminator to evaluate the benefit of direction-specific motion regularization. 

\begin{figure}[!t]
    \vspace{8pt}
    \centering
    \includegraphics[width=1.0\linewidth]{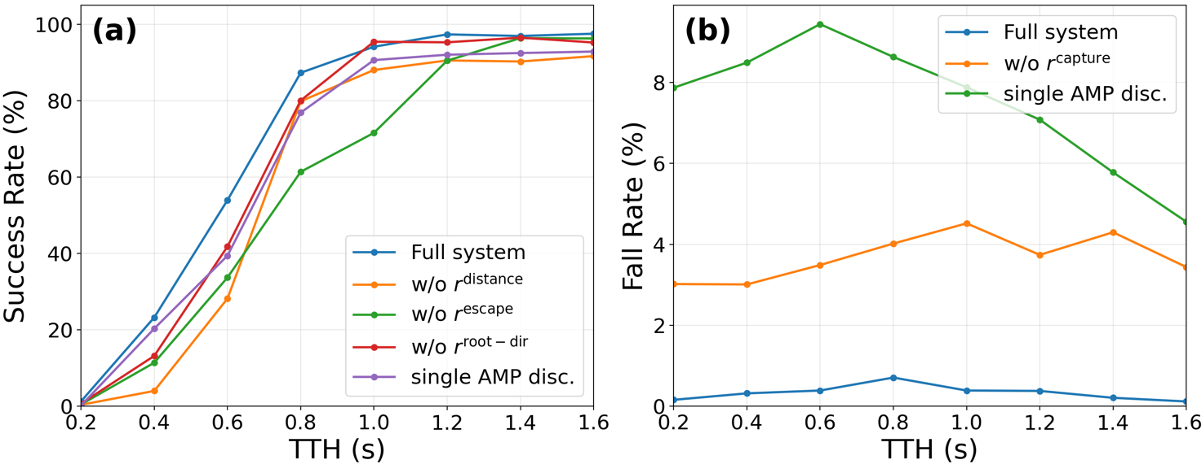}
    \caption{Ablation studies under different time-to-hit ($\mathrm{TTH}$) conditions.
    (a) Success rate after removing individual reward terms or replacing the direction-specific AMP discriminators with a single discriminator.
    (b) Fall rate after removing the capture-point recovery reward or using a single AMP discriminator. }
    \label{fig:ablation}
    \vspace{-1.0em}  
\end{figure}

Fig.~\ref{fig:ablation}(a) evaluates the contribution of the individual reward terms and the multi-discriminator AMP design to obstacle avoidance performance. Removing $r^{\mathrm{distance}}$ consistently reduces the success rate, with the largest degradation observed for short reaction times, indicating that the near-field distance penalty is important for maintaining sufficient clearance from rapidly approaching obstacles. Removing $r^{\mathrm{escape}}$ also leads to a substantial performance drop for $\mathrm{TTH}<1.4$~s, particularly in the more challenging short-$\mathrm{TTH}$ regime, demonstrating that the escape-velocity reward is critical for eliciting timely evasive motions. A similar trend is observed when removing $r^{\mathrm{root-dir}}$: the success rate decreases noticeably for $\mathrm{TTH}<1.0$~s, showing that the root-direction reward helps resolve conflicting local body-part escape directions and promotes coherent whole-body displacement under imminent threats. Replacing the direction-specific AMP discriminators with a single discriminator causes a consistent reduction in success rate, suggesting that direction-specific motion priors improve the effectiveness of the learned evasive behaviors. This result further suggests that promoting a more diverse repertoire of avoidance motions contributes to better obstacle avoidance performance, consistent with the findings of~\cite{xu2025rebot}, which show that greater gait diversity improves dynamic obstacle avoidance.

Fig.~\ref{fig:ablation}(b) further examines the effect of stabilization and motion-prior design on fall rate. Removing $r^{\mathrm{capture}}$ substantially increases the probability of falling across all $\mathrm{TTH}$ ranges, confirming that the capture-point reward is important for recovering balance after highly dynamic evasive maneuvers. Using a single AMP discriminator leads to an even higher fall rate, indicating that direction-specific AMP regularization not only improves avoidance performance but also contributes to more dynamically feasible and stable motions. In contrast, the full system maintains a consistently low fall rate across all $\mathrm{TTH}$ ranges.

\section{Hardware deployment}
We deploy the trained policy on the EngineAI PM01 humanoid to evaluate its performance under real-world conditions. A motion capture system tracks the robot and the incoming obstacle. The obstacle position is transformed into the robot's coordinate frame and provided to the policy. We evaluate obstacle avoidance by throwing a ball with a radius of $0.12$~m toward the robot from different directions and at different speeds. The ball speed recorded by the motion capture system ranges from $3.0$~m/s to $10$~m/s.

\begin{figure*}[h]
    \vspace{8pt}
    \centering
    \includegraphics[width=0.9\linewidth]{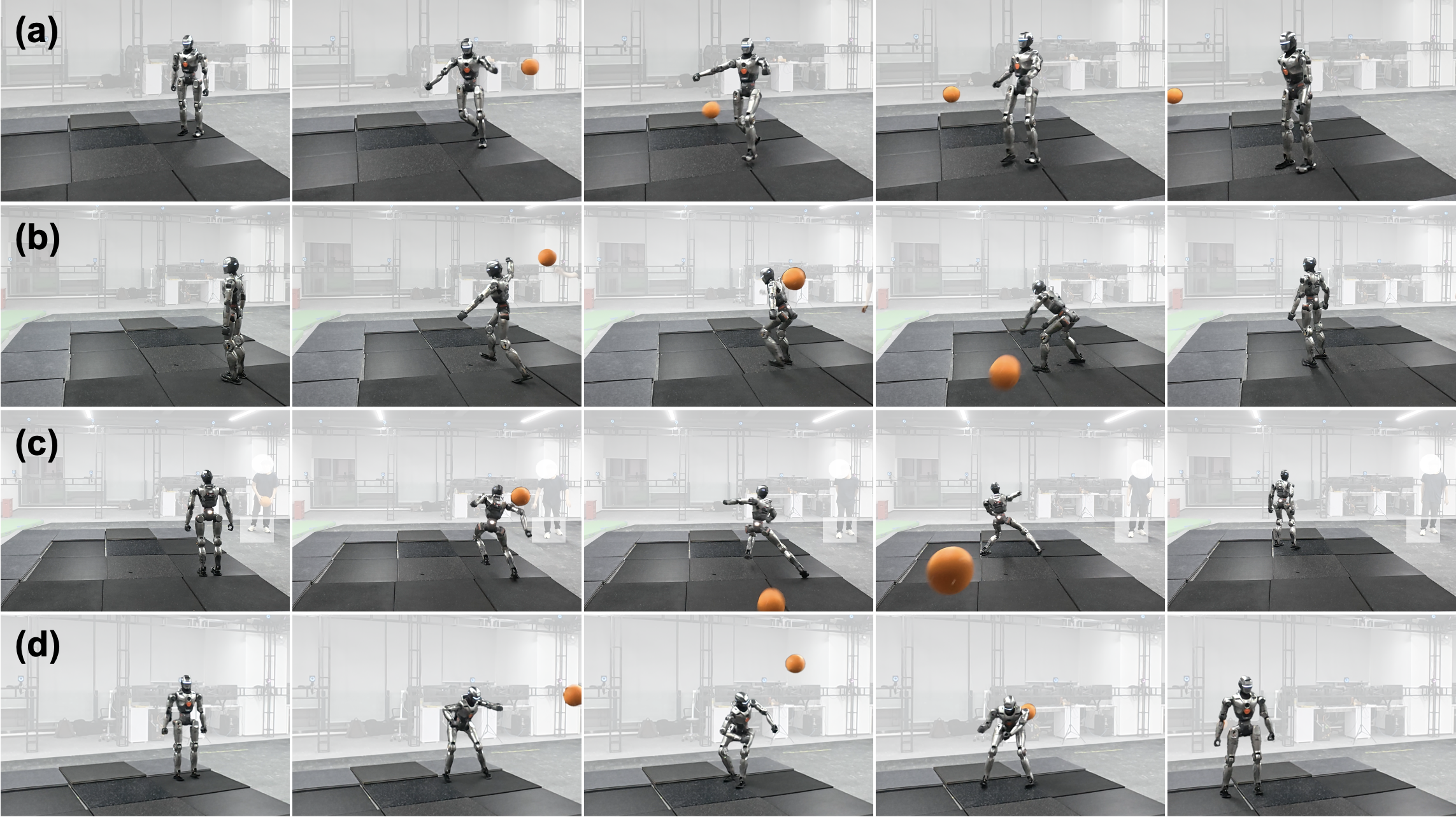}
    \caption{Hardware demonstrations on the EngineAI PM01 humanoid under diverse projectile conditions. (a) A ball aimed toward the robot's center is avoided by stepping forward. (b) A projectile directed toward the head is avoided through a coordinated forward step and ducking motion. (c) A ball aimed at the right shoulder is avoided by stepping laterally to the left. (d) A projectile targeting the head is avoided through a rightward step combined with a ducking maneuver. These examples demonstrate that the learned policy produces diverse whole-body evasive strategies in response to different obstacle trajectories and incoming directions.}
    \label{fig:hardware}
\vspace{-1.0em}   
\end{figure*}

As shown in Fig.~\ref{fig:hardware}, the robot responds with diverse whole-body evasive maneuvers, including forward and lateral steps combined with ducking. Out of the 30 trials, the robot successfully dodges 22 times, achieving an overall success rate of approximately $73\%$. Out of the 8 failure cases, the robot gets hit by the ball 6 times and falls to the ground while dodging twice. Most hit failures occur when fast projectiles (typically \(>8\,\mathrm{m/s}\)) are directed toward the torso; although the threat successfully triggers an evasive response, the remaining reaction time is sometimes insufficient to avoid contact. Another failure mode occurs when a fast-rolling ball approaches the feet: although the robot steps laterally to evade it, the swinging foot inadvertently contacts the ball. These results demonstrate that the learned policy transfers to hardware and generates coordinated evasive behaviors in response to incoming projectiles, with a modest performance gap consistent with expected sim-to-real differences.

\section{Conclusion and future work}
We presented \textit{Humanoid Dodgeball}, a reinforcement learning framework for reactive whole-body avoidance of fast-moving projectiles. By combining threat-aware reward shaping, direction-specific adversarial motion priors, and capture-point guidance, the framework learns diverse evasive maneuvers while promoting balance and post-evasion recovery. Simulation experiments on Unitree G1 and EngineAI PM01 demonstrate higher avoidance success rates than the evaluated baselines across all tested time-to-hit ranges. Ablation studies further support the contributions of the reward design and direction-specific motion priors to avoidance performance and stability. Hardware experiments on PM01 demonstrate coordinated avoidance behaviors.

Despite these results, avoidance remains challenging under short reaction times, and improved success comes at the cost of higher motor power and longer recovery times than PAC-MAN. Future work will explore policies that better balance dodge success rate and power consumption, replace motion capture with onboard perception, and reduce the sim-to-real gap through improved actuator modeling. We also plan to extend the framework to handle multiple obstacles and integrate avoidance with locomotion and manipulation policies.

\bibliographystyle{IEEEtran}
\bibliography{reference.bib}

\end{document}